# Uzbek text summarization based on TF-IDF

**Khabibulla Madatov[1], Shukurla Bekchanov[2], Jernej Vičič[3]**

[1]Urgench state university, 14, Kh. Alimdjan str, Urgench city, 220100, Uzbekistan
habi1972@mail.ru
[2]Urgench state university, 14, Kh. Alimdjan str, Urgench city, 220100, Uzbekistan
shukurla15@gmail.com
[3]Research Centre of the Slovenian Academy of Sciences and Arts, The Fran Ramovš Institute, Novi trg 2, 1000 Ljubljana, Slovenija.  University of Primorska, FAMNIT, Glagoljaska 8, 6000 Koper, Slovenia
jernej.vicic@upr.si

**Abstract**

The volume of information is increasing at an incredible rate with the rapid development of the Internet and electronic information services. Due to time constraints, we don't have the opportunity to read all this information. Even the task of analyzing textual data related to one field requires a lot of work. The text summarization task helps to solve these problems. This article presents an experiment on summarization task for Uzbek language, the methodology was based on text abstracting based on TF-IDF algorithm. Using this density function, semantically important parts of the text are extracted. We summarize the given text by applying the n-gram method to important parts of the whole text. The authors used a specially handcrafted corpus called "School corpus" to evaluate the performance of the proposed method. The results show that the proposed approach is effective in extracting summaries from Uzbek language text and can potentially be used in various applications such as information retrieval and natural language processing. Overall, this research contributes to the growing body of work on text summarization in under-resourced languages.

## 1. Introduction

The task of text summarization is to reduce a given text to a shorter version while retaining the most important information. The Internet, web pages, news, articles, status updates, blogs, as well as the works of scientists who make new discoveries every day, etc., are the source for the immensity of text files. The sheer amount of information generated and distributed in everyday life presents a lot of problems, we have to find, use and process the necessary textual information. In this case, we are faced with the problem of text summarization for text data analysis. Text summarization is one of the most important applications of natural language processing (NLP).

Automatic document summarization has the following preferences:

1) reduces the time of studying the data;
2) speeds up the process of document analysis;
3) facilitates the process of selecting documents when investigating documents;
4) provides more summary options than the traditional text summary.

The minimum requirement for automatic summarization is that the number of text words obtained as a result of automatic summarization does not exceed approximately 30% (Torres-Moreno, 2014, page 33) of the number of original text words.

Automatic text summarization is a complex process, consisting of several phases. The first phase is an initial processing of the text, which often includes the following steps:

- separating the text into parts, sentences, paragraphs, etc.;
- parsing segments into words or tokenization;
- standardization of the words (lemmatization, stemming, etc.);
- removal of the stop words.

Initial processing is a difficult task that largely depends on the language in which the text is written. For example, sentence boundaries are marked by punctuation. Their usage varies considerably from language to language. Also, not all words are separated by spaces in all languages. There are great differences between a text written in an Oriental language such as Chinese, Japanese, or Korean, and a text written in European Latin characters.

This problem also occurs in other steps, such as lemmatization or stemming, the normalization of words using the grammatical part of speech tagging. Even the removal of stop words[1] (X. A. Madatov et al., 2021) depends on the natural language and again it is not a trivial task.

This article is dedicated to Uzbek text summarization based on an algorithm using TF-IDF(Aizawa, 2003). Although the issue of summarization for rich-resourced languages has been solved by various methods, the issue of text summarization for low-resourced languages such as Uzbek still remains an open issue. That is the main motivation for the presented research on text summarization based on TF-IDF. In this case, a dictionary is created from the words contained in the texts and a probability distribution law is created based on their TF-IDF values. In order to assess the results of the proposed method, a corpus named "School corpus" was used, which was created using freely available school books such as "Reading book", "Mother tongue" and "Literature".

---

[1] If the removal of those words from the text not only does not change the context meaning but also leaves the minimum number of words possible that can still hold the meaning of the context, then such words can be called stop words for this work.

**Uzbek language.** The Uzbek language is a Turkic language spoken by more than 30 million people in Uzbekistan and neighboring Central Asian countries. It is the official language of Uzbekistan and is written using the Latin alphabet. Due to its rich history and culture, and the language structure has been heavily influenced by other languages such as Arabic, Persian, and Russian. The Uzbek language has several dialects, with the main dialects being Karakalpak, Khorezm, and Samarkand, among others. Despite its importance, there has been relatively little research on text summarization for the Uzbek language. Our approach aims to fill this gap and provide a useful tool for summarizing Uzbek text.

This paper is structured as follows: We first start with an introduction, in Section 1. Section 2 presents, related works. The main contribution of this work, is presented in Section 3 in the form of methodology. Following, experiments and results are presented in Section 4, which includes information regarding results obtained in this work. Lastly, we conclude this paper with the discussion and conclusion in Section 5.

## 2. Related works

Automatic text summarization originated in the 1950s through the research of Hans Peter Luhn. Luhn first created a model of the summarization of scientific and technical articles(Luhn, 1958). Inspired by his work, many other researchers produced a valuable amount of research output, such as the developed an automatic abstracting system using sentence selection and rejection techniques, and a Word Control List (Edmundson & Wyllys, 1961), which produced high-quality abstracts that warrant large-scale testing (Rush et al., 1971).

Pollock and Zamora studied Chemical Abstracts Service's research on text summarization using a modified Rush-Salvador-Zamora algorithm. They found that some subjects are better suited for automatic extraction and suggest customizing the algorithm for narrow subject areas for better results, they also discussed the viability of automatic extraction (Pollock & Zamora, 1975).

In recent years, researchers have proposed various approaches to improve the effectiveness of text summarization, from cpombining TF-IDF algorithm with the Latent Dirichlet Allocation (LDA) algorithms to generate more informative summaries, to neural network-based approachs that incorporate the attention mechanism to capture the semantic information of the text and improve the summarization quality (Lehnert & Ringle, 2014).

The following works are to the research of the natural language processing of the Uzbek language in (K. Madatov, 2019; K. A. Madatov et al., 2022).

The relationship of Uzbek words is observed in there have been works on gap-filling tasks extracting Uzbek stop words on the example of School corpus is presented in (K. Madatov et al., 2022b) and (K. Madatov et al., 2022c). The level of accuracy of Uzbek stop words detection is presented in (K. Madatov et al., 2022a), Uzbek stemming and lemmatization is discussed in (Sharipov & Sobirov, 2022), are the basic ones used in this research work.

Above all, the Uzbek language has seen a recent grooving trend in the production of NLP-related research works and resources, among them, a machine transliteration tool (Salaev et al., 2022a), sentiment analysis dataset and analyser models (Kuriyozov et al., 2022), as well as semantic evaluation datasets (Salaev et al., 2022b) are some of the many.

## 3. Methodology

The Uzbek language belongs to the family of agglutinative languages, most of the methods for inflected languages text summarizing cannot be directly used for the Uzbek language.

**The scientific novelty of the article:** Creating Uzbek text summarization method based on TF-IDF for the School corpus (K. Madatov et al., 2022c).

In the article (K. Madatov et al., 2022a) considered and proved the problem of finding a part of Uzbek texts, containing stop words for the School corpus. Using this method, it is possible to find the important part of the given text. Below, we present the algorithm of the Uzbek text summarization method. A brief description of steps taken in the implementation of the proposed method is given in Algorithm 1, and the full algorithm can be found in Annex 1.

**Algorithm 1: Uzbek text summarization**

1. The given text (initial text) is separated to words, later addressed as Text
2. Removal of stop words using the dataset extracted from School corpus
3. A unique dictionary from the resulting Text was created, addressed as Text_UW
4. For each $a_i$, which belongs to Text_UW, $w_i=$TF-IDF$(a_i)$ is calculated. Then $w_i$ is transferred $p_i(a)$ by using the formula $p_i=w_i/\Sigma w_i$.
   Density function is created for Text_UW. This step includes the following computational process:(K. Madatov et al., 2022a, 4-5 pages)
   $E=\Sigma i \cdot p_i$ - the mathematical expectation of the unique words
   $D=\Sigma(i-E)^2 \cdot p_i$ - dispersion of the unique words
   $\sigma=\sqrt{D}$ - standard deviation of the unique words
   $E_k=\Sigma p_i \cdot i^k$ -of the unique words
   $\mu_3=E_3-3 \cdot E_1 \cdot E_2+2 \cdot E_1^3$ k -third central moment of the unique words
   $A_s=\mu_3/\sigma^3$ - The asymmetry of the theoretical distribution
5. Important parts of the Text_UW depend on $A_s$. Let $k=E-\sigma$ and $m=E+\sigma$. We find words respectively k- and m-positions Text_UW and position of these words in initial

```
    Text. Let they are k₁ and m₁
    respectively.
 6. a)If Aₛ>0, then the part of the
    Text from the beginning word up
    to k₁-th word part is reloaded
    into the Text, deleting the rest;
    b)If Aₛ<0, then the part of the
    Text from the m₁-th word up to the
    last word is reloaded into the
    Text, deleting the rest.;
    c)If Aₛ=0, then the part of the
    Text from k₁-th word up to m₁-th
    word is reloaded into the Text,
    deleting the rest.
 7. The 3 – gram method is applied to
    the Text
 8. The result is printed.
```

## 4. Experiments and results

The methodology presents the applying of the Uzbek text summarization algorithm to the given text that, solves the complicated problem of finding the significant part and summarization of the given text based on TF-IDF. In this part of the paper we show an example of applying the Uzbek text summarization algorithm.

The adventure novel "Sariq devni minib (Riding the Yellow Giant)" by Kh. Tokhtabayev was chosen as an example for the experiment. A part of this masterpiece is given in the literature textbook of 8th class.

The application of the Uzbek text summarization algorithm to this text can be explained in a step by step recipe as the following:

1. This masterpiece consists of 49,705 tokens and among them 13,740 are unique words.
2. Remove stop words (K. Madatov et al., 2021). As a result, we get Text.
3. Creates a dictionary of Text. This id called unique words and the set of these words are denoted as Text_UW. In this case len(Text_UW)=13740.
4. Each unique word is denoted as $a_i$; $i\epsilon[1...13740]$. For each $a_i$, $w_i$=TF-IDF($a_i$) is calculated. The probability of the word $a_i$ can be calculated using the following formula:

$$p_i=w_i/\sum w_i$$

Numerical values of the probability of unique words are presented in Table 1.

Table -1: Numerical values of the probability of unique words

| Variable | Item |
|---|---|
| E | 4379,22 |
| D | 16019213,55 |
| Σ | 4002,4 |
| σ³ | 64115315874 |
| E₁ | 4379,22 |
| E₂ | 35196780,35 |
| E₃ | 3,40214E+11 |
| μ₃ | 45776660238 |
| Aₛ | 0,714 |

5. K=377.
6. In our case Aₛ>0. The corresponding word for K in Text_UW is "shavla". We find the first occurrence of the word "shavla" in the Text. We cut the part from the first word of the Text to the word "Shavla" and reload to the Text.
7. We apply the 3rd gram to the Text.
8. Print the result. Figure1

The method is presented in Figure 1.

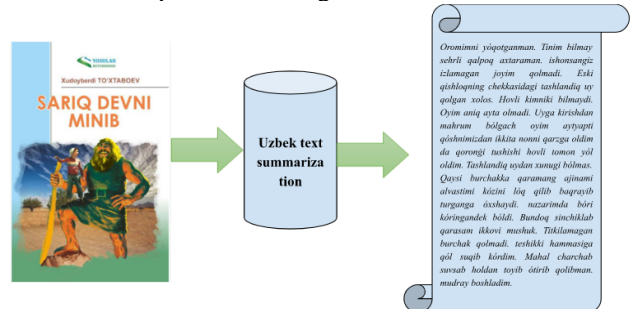

Fig. 1: The result of the application of the Uzbek text summarization algorithm

## 5. Discussion and Conclusion

3 gram were used to summarize the text taken in the experiment. In general, the problem of applying the n-gram to the problem of the Uzbek text summarization and choosing the best n-gram remains open. Despite the openness of this problem, similar problems can be solved using the algorithm, which is the main result of the article. For this, i-gram is taken instead of 3-gram. Using the results obtained for all i-grams, the expert selects the best n-gram.

This article presents an algorithm based on TF-IDF for solving the text summarization for Uzbek language problem. In the process of applying the algorithm, we removed the stop words using the previously created School corpus. The usefulness of the obtained result for further research can be expressed as follows:

1) solving Uzbek natural language processing problems;
2) choosing the best n-gram in the problem of Uzbek text summarization based on TF-IDF;
3) in solving the problem of Uzbek text summarization, how to productive Uzbek text summarization based on TF-IDF in comparison with the trend methods of the Uzbek text summarization. And many other use-cases like these.

**Acknowledgments.** The authors gratefully acknowledge the Erasmus+ program, ELBA (Establishment of training and research centers and courses development on intelligent big data analysis in Central Asia) project (project reference number: 610170-EPP-1-2019-1-ES-EPPKA2-CBHE-JP).

The authors of this paper also present a sincere gratitude to the NLP team at Urgench State University for the enormous support in discussions during the model creation and experiments.

# Annex

Annex 1: The full algorithm of a TF-IDF based text summarization for Uzbek texts.

*1:INPUT: (Text)*
*2:     Token(Text)*
*3:          Remove(Collacation_Two_Words(Text))*
*4:          Remove(Unigram(Text))*
*5:          Remove(RuleBase(Text))*
          *//{pronoun, adverb, conjunction, introductory word, adverbial word, auxiliary word, prepositions}*
*6:          TF-IDF(Text)*
*7:          p(Text_UW)= TFIDF(Text_UW)/SUM(TFIDF(Text_UW))*
          *// Text_unique words- Text_UW*
*8:          E=SUM(Text_UW)*p(Text_UW)*
*9:          D=SUM(Text_UW -E)^2*p(Text_UW)*
*10:         SIGMA=SQR(D)*
*11:         EI(m)=SUM(p(Text_UW))* Text_UW ^m// m – 1..len(Text_UW)*
*12:         M1=E*
*13:         M2= SUM(p(Text_UW))* Text_UW ^2*
*14:         M3= SUM(p(Text_UW))* Text_UW ^3*
*15:         MIYU=M3-3*M1*M2+2*M1^3*
*16:         AS= MIYU/ SIGMA^3*
*17:     If(AS>0) then Text2=TFIDF[0;E-SIGMA]*
*18:     If(AS<0) then Text2=TFIDF[E-SIGMA- len(Text_UW)]*
*19:     If(AS=0) then Text2=TFIDF[E-SIGMA; E+SIGMA]*
*20:     Gram3(Text2)*
*21:          Token(Text2)*
*22:          SummarText:= P(token(i)|token(i+1)| token(i+2))*
*23:     Print(SummarText)*
*24:END*